\pdfoutput=1
\documentclass[lettersize,journal]{IEEEtran}
\usepackage{amsmath,amsfonts}
\usepackage[ruled,linesnumbered]{algorithm2e}
\usepackage{array}
\usepackage[caption=false,font=normalsize,labelfont=sf,textfont=sf]{subfig}
\usepackage{textcomp}
\usepackage{stfloats}
\usepackage{bm}
\usepackage{url}
\usepackage{verbatim}
\usepackage{graphicx}
\usepackage{diagbox}
\usepackage{cite}
\usepackage{multirow}
\usepackage[table,xcdraw]{xcolor}
\usepackage{threeparttable}
\usepackage[implicit=false]{hyperref}
\usepackage{hyperref}
\hypersetup{hidelinks,
	colorlinks=true,
	allcolors=black,
	pdfstartview=Fit,
	breaklinks=true}

\newcommand{\orcid}[1]{\href{https://orcid.org/#1}{\includegraphics[width=10pt]{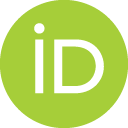}}}

\makeatletter
\def\bstctlcite{\@ifnextchar[{\@bstctlcite}{\@bstctlcite[@auxout]}}
\def\@bstctlcite[#1]#2{\@bsphack
  \@for\@citeb:=#2\do{%
    \edef\@citeb{\expandafter\@firstofone\@citeb}%
    \if@filesw\immediate\write\csname #1\endcsname{\string\citation{\@citeb}}\fi}%
  \@esphack}
\makeatother

\begin{document}

\bstctlcite{BSTcontrol}

\title{
FireFly v2: Advancing Hardware Support for High-Performance Spiking Neural Network with a Spatiotemporal FPGA Accelerator
}

\author{
    Jindong Li \orcid{0000-0002-4009-916X},
    Guobin Shen \orcid{0000-0002-4069-2107},
    Dongcheng Zhao \orcid{0000-0002-0593-8650}, 
    Qian Zhang \orcid{0000-0001-5314-4233},
    Yi Zeng \orcid{0000-0002-9595-9091}

\thanks{Manuscript created 20 September 2023. This work was supported by the Chinese Academy of Sciences Foundation Frontier Scientific Research Program (ZDBS-LY- JSC013). \textit{(Corresponding authors: Qian Zhang; Yi Zeng.)}}

\thanks{Jindong Li and Qian Zhang are with the School of Artificial Intelligence, University of Chinese Academy of Sciences, Beijing 100049, China, and also with the Brain-inspired Cognitive Intelligence Lab, Institute of Automation, Chinese Academy of Sciences, Beijing 100190, China (e-mail: lijindong2022@ia.ac.cn, q.zhang@ia.ac.cn).}

\thanks{Guobin Shen is with the School of Future Technology, University of Chinese Academy of Sciences, Beijing 100049, China, and also with the Brain-inspired Cognitive Intelligence Lab,  Institute of Automation, Chinese Academy of Sciences, Beijing 100190, China (e-mail: shenguobin2021@ia.ac.cn).}

\thanks{Dongcheng Zhao is with the Brain-inspired Cognitive Intelligence Lab, Institute of Automation, Chinese Academy of Sciences, Beijing 100190, China (e-mail: zhaodongcheng2016@ia.ac.cn).}  

\thanks{Yi Zeng is with the Brain-inspired Cognitive Intelligence Lab, Institute of Automation, Chinese Academy of Sciences, Beijing 100190, China, and University of Chinese Academy of Sciences, Beijing 100049, China, and Center for Excellence in Brain Science and Intelligence Technology, Chinese Academy of Sciences, Shanghai 200031, China (e-mail: yi.zeng@ia.ac.cn).}
}

\maketitle

\begin{abstract}
Spiking Neural Networks (SNNs) are expected to be a promising alternative to Artificial Neural Networks (ANNs) due to their strong biological interpretability and high energy efficiency.
Specialized SNN hardware offers clear advantages over general-purpose devices in terms of power and performance.
However, there's still room to advance hardware support for state-of-the-art (SOTA) SNN algorithms and improve computation and memory efficiency.
As a further step in supporting high-performance SNNs on specialized hardware, we introduce FireFly v2, an FPGA SNN accelerator that can address the issue of non-spike operation in current SOTA SNN algorithms, which presents an obstacle in the end-to-end deployment onto existing SNN hardware.
To more effectively align with the SNN characteristics, we design a spatiotemporal dataflow that allows four dimensions of parallelism and eliminates the need for membrane potential storage, enabling on-the-fly spike processing and spike generation.
To further improve hardware acceleration performance, we develop a high-performance spike computing engine as a backend based on a systolic array operating at 500-600MHz.
To the best of our knowledge, FireFly v2 achieves the highest clock frequency among all FPGA-based implementations. Furthermore, it stands as the first SNN accelerator capable of supporting non-spike operations, which are commonly used in advanced SNN algorithms.
FireFly v2 has doubled the throughput and DSP efficiency when compared to our previous version of FireFly and it exhibits $\times 1.33$ the DSP efficiency and $\times 1.42$ the power efficiency compared to the current most advanced FPGA accelerators.
\end{abstract}

\begin{IEEEkeywords}
Spiking Neural Networks, Field-programmable gate array, Hardware Accelerator, Non-Spike Operation, Spatiotemporal Dataflow
\end{IEEEkeywords}

\section{Introduction}
\IEEEPARstart{S}{piking} neural networks (SNNs) are considered a promising alternative to artificial neural networks (ANNs) due to their high biological plausibility, event-driven nature, and low power consumption\cite{maass1997networks}. Recent advancements in SNN algorithms have drawn inspiration from both biological evidence and deep learning insights, narrowing the performance gap with ANNs\cite{shen2022backpropagation}\cite{zheng2021going}. However, current neuromorphic hardware\cite{davies2018loihi}\cite{akopyan2015truenorth}\cite{painkras2013spinnaker} cannot match the performance of ANN accelerator counterparts and, worse still, cannot support state-of-the-art SNN algorithms.

\begin{figure}
  \centering
  \includegraphics[width=1.0\linewidth]{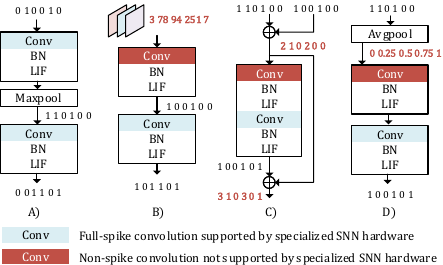}
  \caption{A) An SNN backbone with full-spike operation. B) The non-spike convoluton in the direct coding layer.  C) The sum of spikes in SEW-ResNet introduces non-spike convolution. D) The average pooling layer introduces fractional-spike convolution.}
  \label{fig:algo}
\end{figure}

Ongoing research aimed at developing high-performance SNN algorithms has made significant strides in narrowing the benchmark accuracy gap between ANNs. However, it also presents challenges for specialized SNN hardware due to the introduction of hardware-unfriendly computation, shown in Fig.\ref{fig:algo}.
Current SNN algorithms typically use direct input encoding\cite{rathi2021diet} with analog pixel values applied to the initial convolutional layer, followed by spiking neurons for end-to-end backpropagation, improved benchmark accuracy and reduced time steps. However, the direct encoding convolutional layer poses compatibility challenges with existing specialized SNN hardware designed for spike-based computation.
Current deep SNN models incorporate a residual connection by applying spike-element-wise summation between the residual path and the shortcut path\cite{fang2021deep}. However, the sum of spikes operations introduce non-spike operation in the next convolutional layer.
Furthermore, the commonly employed Average Pooling function in SNN models introduces fractional-spike convolution, which is not supported by current SNN hardware equipped with spike-based computing engines.
Existing SNN software frameworks for deployment, such as Lava\cite{lavaweb} for Loihi\cite{davies2018loihi}, are unable to address these issues since neuromorphic hardware inherently lacks support for non-spike operations.

Advancements in current specialized SNN hardware continue to pursue low power and high performance through architectural designs. However, there is still room for improvement in terms of spatiotemporal dataflow, parallelism scheme and computing engine design, particularly in field-programmable gate array (FPGA) implementations.
Current FPGA SNN accelerators either have no parallelism in the temporal domain\cite{chen2022cerebron} or sacrifice spatial parallelism for temporal parallelism\cite{lee2020reconfigurable}.
Moreover, systolic-array-based SNN accelerators still have naive implementations of spiking computing engines\cite{wang2020sies} that run at low frequencies and have limited parallelism.
Our previous version of FireFly\cite{li2023firefly} employed a high-performance systolic array running at 300MHz with DSP optimizations, but it still had limited parallelism dimensions and ran at a frequency far from the theoretical extreme frequency on Xilinx Ultrascale FPGA.
FireFly adopted a weight-stationary dataflow and designed a synaptic weight delivery hierarchy to enable efficient weight data reuse, but it still required large on-chip membrane potential storage and did not support temporal parallelism.

As agile development methodologies for customized hardware evolve, the gap between the development cycle of customized hardware and the iteration speed of algorithms is gradually closing. We acknowledge the importance of aligning research on SNN hardware accelerators with the advancements in SNN algorithms.
In this work, we introduce FireFly v2 as another step to advance specialized hardware support for SOTA SNN algorithms while further enhancing hardware performance. FireFly v2 brings several significant improvements:

1) FireFly v2 is a general FPGA SNN accelerator that can support 
A) non-spike operations in direct input encoding\cite{rathi2021diet} and spike-element-wise ResNet\cite{fang2021deep}. 
B) multiple neurodynamics, such as IF\cite{abbott1999lapicque}, LIF\cite{dayan2003theoretical} and RMP\cite{han2020rmp} neurons. 
C) arbitrary convolutional configurations such as different kernel sizes, strides, and pads. These integrations cover many recent SNN advancements.

2) FireFly v2 utilizes a spatiotemporal dataflow scheme for SNNs that enables four dimensions of parallelism. FireFly v2 not only process firing neurons on the fly but also generate spikes on the fly, eliminating the need for expensive membrane potential storage thus greatly reducing on-chip memory consumption and inference latency compared to the serial processing of spikes at each time step. 

3) FireFly v2 integrates a high-performance spiking computing engine with a systolic array that supports four dimensions of parallelism and runs at 500-600MHz on different FPGA devices, which is closer to the extreme clock frequency of Xilinx Ultrascale FPGA and thus achieves $\times 1.67-\times 2$ throughput and DSP efficiency compared to the original FireFly\cite{li2023firefly}. Compared with the existing most advanced SNN accelerator DeepFire2, FireFly v2 achieves $\times 1.33$ the DSP efficiency and $\times 1.42$ power efficiency on a much smaller FPGA edge device.

The remaining sections of the paper are organized as follows:
Section II presents related work that shares motivation with our research.
Section III introduces how we address the non-spike operation challenge.
Section IV describes our proposed spatiotemporal dataflow.
Section V outlines the hardware architecture of FireFly v2, including the design of the 500-600MHz spike computing engine.
Section VI provides details on experiments regarding hardware specifications and benchmark evaluations.
Finally, Section VII concludes the paper.

\section{Related Work}

Rather than attempting to cover all neuromorphic hardware or SNN accelerators relevant to our research, we will focus on studies that share a similar motivation to our own and explore potential improvements to these works.

\subsection{Supporting Versatile SNNs with a Single Hardware Engine}

Using a single computation engine has emerged as a favored design option for FPGA-based neural network accelerators, allowing for the deployment of a variety of models without requiring fabric reconfiguration.
While ANN variants primarily differ in convolutional configurations and structural designs, SNN variants are much more varied and complex as they also differ in input encoding schemes and neuron types. However, only a limited amount of research has explored the design of a unified SNN accelerator. Cerebon\cite{chen2022cerebron} designed a reconfigurable compute engine compatible with a variety of spiking convolutional layers including depthwise and pointwise convolutions. Ye et al.\cite{ye2022implementation} designed a neuromorphic platform that supports SNNs with MLP and CNN Topologies. Zhang et al. \cite{zhang2022configurable} proposed an architecture that supports multiple coding schemes including rate coding and temporal coding. However, these studies combined only address a small fraction of the many SNN variants and did not cover the recent SNN advancements.

In this paper, our objective is to narrow the gap between modern SNN algorithms and hardware accelerators. We achieve this by supporting general non-spike operations, including direct encoding, spike-element-wise residual connections, and the frequently used Average Pooling operation. Additionally, our approach supports various convolutional configurations in a dynamically reconfigurable fashion and offers versatility in neuron types through static reconfiguration.

\subsection{Dataflow and Parallelism Schemes for SNNs}

Existing SNN accelerators with dataflow and parallelism schemes exploration have limited parallelism dimensions. 
Lee et al. \cite{lee2020reconfigurable} proposed a Psum-friendly dataflow and a 2D systolic array that enables spatiotemporal parallelism. However, they only support output channel parallelism in the spatial domain, which sacrifices spatial parallelism for temporal parallelism.
Lien et al. \cite{lien2022sparse} investigated the convolution dataflow and identified a specific loop order that suits its spatial parallelism scheme, or pixel-level parallelism scheme using block convolution. However, the proposed method does not support temporal parallelism.
Spinalflow \cite{narayanan2020spinalflow} presented a specialized architecture and dataflow for SNNs that exclusively supports temporal coding, where neurons only fire once in all time steps. It sorts the synaptic input spikes chronologically and can handle a maximum of 2048 non-zero spikes within the receptive field, processing spikes sequentially. It updates 128 neurons from different channels one spike packet at a time, which means that it only supports output channel parallelism.
SATO \cite{liu2022sato} expands the neuron-level parallelism to additional temporal-level parallelism by parallelizing the integration of received spikes at each time step. SATO also achieved impressive sparsity acceleration and supported workload balancing. However, similar to Spinalflow, SATO only supports temporal-coded SNNs.

In this paper, we propose a spatiotemporal dataflow with four dimensions of parallelism, namely input channel parallelism, output channel parallelism, pixel-level parallelism and time step parallelism.

\subsection{FPGA Accelerator with DSP Optimization Techniques}

DSP optimization techniques are commonly used in FPGA-based computational-intensive accelerator designs. Here, we focus on DSP48E2 optimization techniques in Xilinx Ultrascale FPGAs.
Previous ANN accelerators have demonstrated the ability to fully utilize the capabilities that DSP slices provide.
Xilinx's white paper\cite{fu2016deep} documents that the $27\times 18$ multiplier in DSP48E2 can be split into two $8\times 8$ multipliers with a shared input operand.
Recent studies\cite{wu2017high} in ANN accelerators have proposed the concept of DSP supertiles by storing operands in distributed RAM around the DSP48E2, which allows for reaching the extreme clock frequency of Xilinx Ultrascale FPGAs.
Vitis AI has adopted the DSP double data clock technique in its DPU design. This allows the systolic array to run at double the clock frequency while the rest of the system runs at the base frequency, achieving high inference performance for FPGA platforms.

Nonetheless, these optimization techniques are tailored for multiplication-intensive ANN accelerator designs and may not be directly applicable to the SNN computation workload. The effectiveness of DSP48E2 in accelerating SNN computation is not immediately apparent.
DeepFire\cite{aung2021deepfire} was the first research to utilize the SIMD feature of DSP48E2s in the SNN accelerator domain. They implemented the AND operations between spikes and synaptic weights using the LUT fabric and used one DSP48E2 and three fabric adders to build an 8-input synaptic currents integration circuit. DeepFire2\cite{aung2023deepfire2} recognized the inefficiency of implementing a 2-input AND operation using a 6-input LUT and avoided AND operations by considering the spike as the synchronous reset of the flip-flop register. This further improved the system frequency but was still not the most ideal implementation. Our previous version of FireFly\cite{li2023firefly} proposed the most efficient implementation of synaptic operation with a fabric-free approach. We utilized a wide-bus multiplexer, SIMD adder, and dedicated cascade path inside the DSP48E2 to construct a $16\times 4$ synaptic crossbar using eight cascaded DSP48E2 slices. This cascaded chain was used in FireFly to build a large systolic-array-based spiking computing engine. However, fabric circuits of other system components limit the frequency in a single clock system.

In this paper, we continue to adopt the fabric-free implementation of synaptic operation proposed in our previous version of FireFly but with a different dataflow. We further incorporate DSP double data rate techniques adopted by the Vitis AI DPU. As a result, FireFly v2 achieves significant speed up compared to its previous version.

\section{Addressing the Non-Spike Operation Challenge}

In recent SNN advancements, the inclusion of non-spike operations has been proven to be a challenging task for specialized SNN hardware. The primary challenge lies in achieving concurrent support for both spike and non-spike operations while maintaining an acceptable level of hardware overhead. Remarkably, none of the existing specialized SNN hardware has undertaken this challenge or even considered its feasibility.
In FireFly v2, we present a solution for tackling the challenge posed by non-spike operations. Without loss of generality, we consider three typical non-spike operation scenarios: pixel operation in direct encoding, multi-bit spike operation in SEW ResNet, and fractional spike operation introduced by average pooling.

Instead of designing separate processing units for these different non-spike operation cases, FireFly v2 draws inspiration from the field of bit-serial accelerators to address the non-spike operation challenge.
In bit-serial accelerators\cite{ryu2022bitblade}\cite{sharma2018bit}, operands are divided into smaller bits, computations are carried out using low-bit arithmetic logic in the processing unit, and the partial sums are shifted and merged to reconstruct full-precision results.
In FireFly v2, we utilize a single spike computing engine to perform spike-weight multiplications, decompose the non-spike operands and flexibly merge the partial sums to support non-spike operations. We focus on the three aforementioned common scenarios of non-spike operations.

\paragraph{Pixel Convolution in Direct Encoding Layer} In the case of 8-bit pixel convolution using direct input encoding, we treat the 8-bit pixel as spikes occurring over 8 equivalent time steps. We then perform spike-weight convolution for each time step and combine the resulting 8 partial sums using shift-add logic.
\paragraph{Multi-bit Spike Convolution in SEW-ResNet} In a certain SEW ResNet layer, a $B$-bit spike in $T$ simulation time steps can be deconstructed into a sequence of binary spikes spanning $B\times T$ equivalent time steps. Partial sums are then shifted and merged $B$ in a group to reconstruct the actual partial sums of $T$ actual time steps. 
As the sum of spikes accumulates, SEW-ResNet will produce $log(N+1)$-bit spikes after $N$ spike-element-wise residual connections. With a 4-bit spike representation, this accommodates up to 16 residual connections, aligning with the SEW-ResNet34 architecture.
\paragraph{Fractional Spike Convolution introduced by AveragePooling} The fractional values can be left-shifted to integers, treated as multi-bit spikes, and the partial sums can be right-shifted by the same amount to obtain the accurate partial sum. For example, consider a $2\times 2$ average pooling operation that yields values such as $0, 0.25, 0.5, 0.75, 1$. These values can be left-shifted by 2 positions, resulting in $0, 1, 2, 3, 4$, and treated as 3-bit spikes.

\begin{figure}
  \centering
  \includegraphics[width=1.0\linewidth]{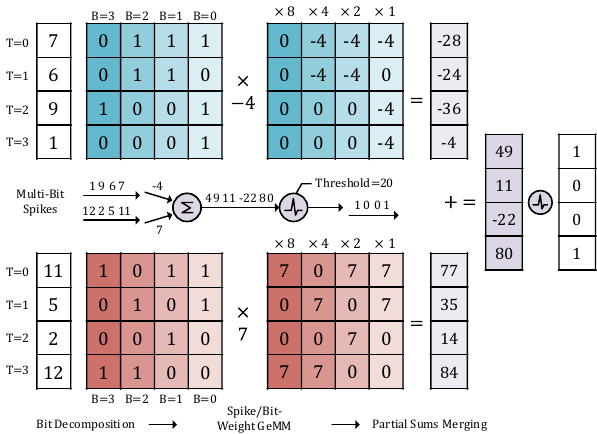}
  \caption{Addressing the computation of multi-bit spikes through bit decomposition, bit-weighted GeMM, and partial sums merging. The figure illustrates a case of 4-bit spikes computation.}
  \label{fig:solution}
\end{figure}

Fig. \ref{fig:solution} illustrates the computation flow for 4-bit spikes with $T=4$. In this scenario, the spiking neuron receives two 4-bit spike inputs from synapses and generates binary spikes. The 4-bit spike sequence is initially decomposed bit by bit, creating a $4\times 4$ bit matrix. Each bit is then multiplied with the corresponding synaptic weight, and the partial sums of each bit are further scaled by factors of 8, 4, 2, and 1 (or left-shifted by 3, 2, 1, and 0), respectively. These scaled values are then summed together, resulting in the actual synaptic current at each time step. Subsequently, these synaptic currents are accumulated and reset when they reach a certain threshold, leading to the generation of spikes according to specific neurodynamic behaviors.

Although the bit-serial decomposition of the multi-bit spikes can fully support the non-spike operations existing in state-of-the-art SNN algorithms without any accuracy drop, it inevitably leads to an increased computational workload. 
After three cascaded residual connections in SEW ResNet, the maximum value of spikes reaches $4$, necessitating a 3-bit representation.
Similarly, spikes with fractional values yielded by the most commonly seen $2\times 2$ average pooling also need a minimum 3-bit representation.
To prevent the potential escalation of spike bit-width as the SNN network increases in depth, we introduce a saturate-or-shift approach to confine the bit-width of spikes within 2 bits.
In our approach, an analysis of data distribution will be conducted in advance on a batch of representative samples in software, similar to the post-training quantization. This analysis will guide the decision of whether to saturate the spike value exceeding $4$ to $3$, or perform a right shift operation on all spike values, resulting in the range of $0, 1, 2$, accompanied by a corresponding left shift of the partial sum to recover the results.
This method ensures that the bit-width of spikes remains confined within 2 bits, effectively mitigating the escalation of the computational workload.

\section{Optimized Spatiotemporal Dataflow}

\begin{figure*}
  \centering
  \includegraphics[width=1.0\linewidth]{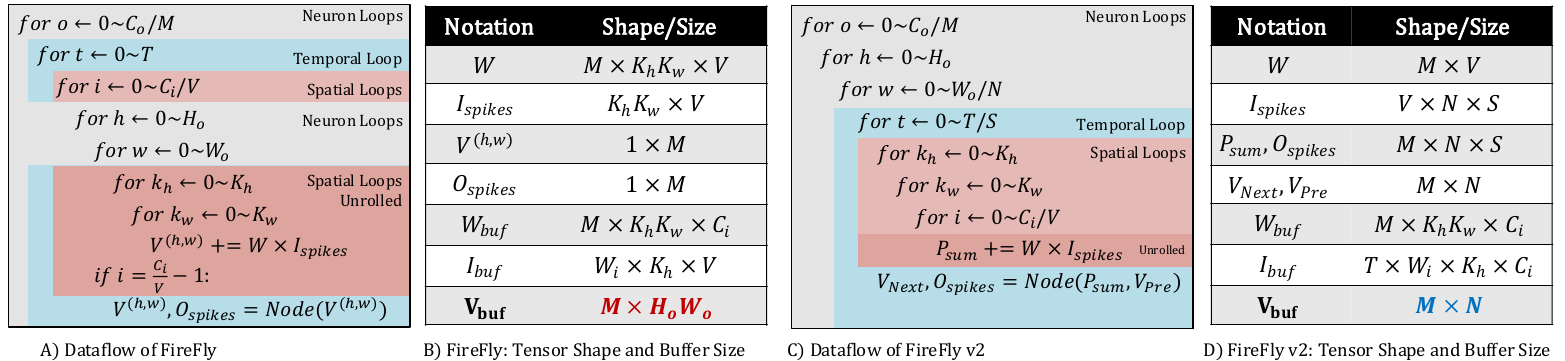}
  \caption{
  Comparison of the dataflow in FireFly v2 to its previous version FireFly. A) FireFly's dataflow design. B) Mentioned tensor shape or buffer size in FireFly's dataflow. C) FireFly v2's dataflow design. D) Mentioned tensor shape or buffer size in FireFly v2's dataflow.
  }
  \label{fig:dataflow}
\end{figure*}

\begin{table}[]
  \begin{center}
    \caption{Notations}
    \label{note}
  \begin{tabular}{c|c}
  \hline
  Notation & Description                        \\ \hline\hline
  $T$        & Time Step           \\ \hline
  $H_o,W_o$        & Size of Output Feature Map \\ \hline
  $H_i,W_i$        & Size of Input Feature Map  \\ \hline
  $K_h,K_w$        & Kernel Size                \\ \hline
  $C_o,C_i$        & Input Output Channels      \\ \hline\hline
  $M$        & Output Channel Parallelism \\ \hline
  $V$        & Input Channel Parallelism  \\ \hline
  $N$        & Pixel Parallelism          \\ \hline
  $S$        & Time Step Parallelism      \\ \hline
  \end{tabular}
  \end{center}
\end{table}

Although the dataflow of ANNs has undergone extensive study, achieving the balance between the spatial and temporal dimensions in the dataflow of SNNs proves to be a challenging task.
In FireFly v2, we propose a spatiotemporal dataflow that builds upon the output stationary dataflow while incorporating variable tiling and parallelism schemes designed specifically for SNNs. In contrast to the previous version of FireFly\cite{li2023firefly}, our approach significantly reduces memory consumption and enables a higher degree of parallelism and reconfigurability. 

We focus on the dataflow of a single convolution layer. Although FireFly v2 does support non-spike convolution, we focus on the 1-bit spike case in this section to simplify the dataflow illustration.
We first explain the variables used in this paper, which are also listed in Table.\ref{note}.
$T$ represents the total number of time steps. $H_o$ and $W_o$ represent the height and width of the output feature map, respectively, while $H_i$ and $W_i$ represent the height and width of the input feature map, respectively. $K_h$ and $K_w$ denote the height and width of the kernel, while $C_i$ and $C_o$ represent the input and output channels, respectively. We leverage four dimensions of parallelism and perform variable tiling on the output channel $C_o$, input channel $C_i$, the width of the output feature map $W_o$, and equivalent time step $T_e$. We denote the output channel parallelism as $M$, input channel parallelism as $V$, pixel parallelism as $N$ and time step parallelism as $S$.

Similar to convolution in ANNs, convolution in SNNs can be expressed using nested for-loops, accompanied by an additional time-step loop. The permutation of loop order does not alter the computation results, yet it does influence data locality and data reuse opportunities.
To simplify the illustration, we employ an ordered tuple to represent the permutation of folded or unrolled for-loops, with the unrolled for-loops enclosed in square brackets, denoted as $[]$.
The fully folded computation flow of an SNN convolution layer can be represented as $(T, C_o, H_o, W_o, C_i, K_h, K_w)$. 
Here, we employ a similar loop notation as presented in SATO\cite{liu2022sato}.
The dimensions $(C_o, H_o, W_o)$ correspond to the neuron loops, representing independent neurons within a convolutional layer, indicated by the color blue in Fig.\ref{fig:dataflow}. On the other hand, the dimensions $(C_i, K_h, W_h)$ represent the spatial loops, denoting spatially fan-in neurons and are marked with the color orange.

\paragraph{Inefficient Dataflow in FireFly} In the previous version of FireFly, we parallelled the computation for both the input-output channel dimension and the kernel dimension. This resulted in a dataflow represented as $(\frac{C_o}{M}, T, \frac{C_i}{V}, H_o, W_o, [M, K_h, K_w, V])$, as depicted in Fig. \ref{fig:dataflow}A.
The spike computing engine within FireFly conducts matrix-vector multiplication between the $K_h K_w\times V$ binary spikes and the $M\times K_h K_w \times V$ synaptic weight matrix, yielding a $1\times M$ partial sum. Subsequently, when the last fragment of spikes from the tiled input channel passes through, the neurodynamics calculation is performed to generate a $1\times M$ output spike vector.
To achieve weight data reuse across $T$ time steps, an on-chip cache is necessary to store $M\times K_h K_w \times C_i$ synaptic weights. To avoid the need for off-chip storage and loading of multi-bit membrane potential, it becomes imperative to temporarily store $M\times H_o W_o$ membrane potential values on-chip. However, this poses potential issues, particularly when dealing with large feature maps. The aforementioned tensor shapes or buffer sizes in FireFly's architecture are denoted in Fig.\ref{fig:dataflow}B.

\paragraph{Spatiotemporal Dataflow in FireFly v2} In FireFly v2, we tackle the following limitations in the FireFly architecture's dataflow:
1) The need for large on-chip storage of membrane potential. 2)The constraint of a fixed convolution configuration resulting from parallelism on the kernel dimension. 3) The absence of parallelism at both the temporal and pixel levels. The adopted spatiotemporal dataflow scheme is depicted in Fig.\ref{fig:dataflow}C.
To address the long data dependencies spanning $C_o\times H_o\times W_o$ neurons across $T$ time steps, we've rearranged the loops by placing the temporal loop after the neuron loops and before the spatial loops. We have also tiled on both the width of the output feature map $W_o$, and the time step $T$, to introduce two additional dimensions of parallelism. After the reordering and tiling, the resulting dataflow takes the form of $(\frac{C_o}{M}, H_o, \frac{W_o}{N}, \frac{T}{S}, K_h, K_w, \frac{C_i}{V},[M, V, N, S])$, shown in Fig.\ref{fig:dataflow}C. The spike computing engine performs matrix multiplication between the $V\times N\times S$ binary spike matrix and the $M\times V$ weight matrix, yielding a $M\times N\times S$ partial sum matrix. After accumulating the fan-in pre-synaptic currents across the $K_h, K_w, \frac{C_i}{V}$ spatial loops, neurodynamic calculations are carried out by incorporating the $M\times N\times S$ partial sums along with the residual $M\times N$ membrane potential $V_{Pre}$ from the previous time step batch. This process results in the generation of $M\times V\times S$ output spikes and the next residual membrane potentials $V_{Next}$. The aforementioned tensor shapes or buffer sizes in FireFly v2's architecture are denoted in Fig.\ref{fig:dataflow}D.

This spatiotemporal dataflow framework enables four dimensions of parallelism including output channel parallelism, input channel parallelism, pixel-level parallelism, and time step parallelism.
Through the separation of the parallelism scheme from the kernel dimension, we enable support for various convolution configurations, accommodating differences in kernel size and stride.
By positioning the temporal loop as the innermost loop before the spatial loops, the necessity to cache only $M\times N$ residual membrane potentials on-chip becomes inconsequential.
Without the explicit storage of the membrane potentials, spikes are not only processed but also generated on the fly. The only additional overhead is the need for increased storage space for input spikes. Given that input spikes require much less memory storage compared to membrane potentials, this added requirement is of minimal concern.

\section{Hardware Architecture}

\begin{figure*}
  \centering
  \includegraphics[width=1.0\linewidth]{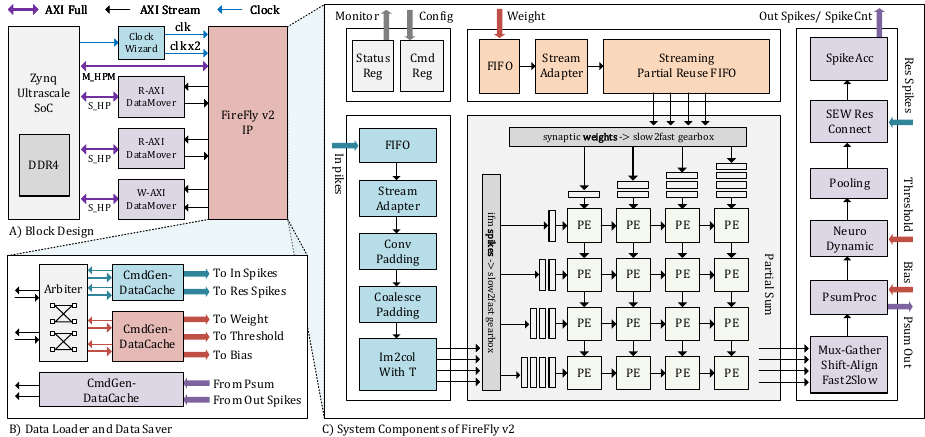}
  \caption{Hardware Architecture of FireFly v2. A) The system block design of FireFly v2. B) The efficient data loader and data saver of FireFly v2. C) The key system components of FireFly v2. The blue blocks represent a series of input spike preprocessing modules, whereas the orange blocks signify a sequence of synaptic weight preprocessing modules. The purple blocks represent the post-processing modules for the partial sums and output spikes.}
  \label{fig:bd}
\end{figure*}

The hardware architecture of FireFly v2 is illustrated in Fig.\ref{fig:bd}. Fig.\ref{fig:bd}A provides an overview of FireFly v2's system block diagram.
In the customized PL system of the Zynq Ultrascale SoC, a clock wizard IP is instantiated to generate two synchronous clocks, with one clock operating at twice the frequency of the other.
One master M-AXI-HPM port of the Zynq Ultrascale SoC is utilized for command configuration status control, directly connected to the FireFly v2 IP. Two 128-bit S-AXI-HP ports are enabled and are connected to two read-only AXI DataMovers respectively, facilitating high-speed PS to PL data transfer. Additionally, another 128-bit S-AXI-HP port is employed and connected to a write-only AXI DataMover, enabling PL to PS data transfer. The FireFly v2 IP interfaces with all three AXI DataMovers.
Please note that we have not utilized all of the S-AXI-HP ports available on the Zynq SoC, and the peak memory bandwidth of the three instantiated S-AXI-HP ports is far from the Zynq Ultrascale device's limit of 19.2 GB/s.

We have designed an efficient data loader and data saver module to fully harness the bandwidth capacity of the three S-AXI-HP ports, as illustrated in Fig.\ref{fig:bd}B.
We have instantiated two CmdGen-DataCache Units to generate Address-Length data transfer commands for input or residual spikes, as well as parameters such as weights, bias, and thresholds, respectively. Additionally, another CmdGen-DataCache is dedicated to handling output spikes or partial sums transfer, utilizing the write-only AXI DataMover interface.
Given the availability of two read-only AXI DataMover interfaces, the simplest approach would be to assign each CmdGen-DataCache unit to occupy one AXI DataMover interface. However, in various convolutional configurations, the required data bandwidth for input spikes and synaptic weights differs. For instance, in a $1\times 1$ convolution, the synaptic weights demand relatively less bandwidth, while input spikes require more. Conversely, in a $3\times 3$ convolution with small $4\times 4$ images, the synaptic weights require greater bandwidth, while the input spikes need less.
To efficiently utilize the total available bandwidth and prevent one AXI DataMover from being fully utilized while the other remains idle, we have introduced an arbiter. This arbiter arbitrates incoming commands from the CmdGen-DataCache units and the data flow from the AXI DataMover interfaces, enabling flexible bandwidth balancing for varying spikes and weight workloads.

The key system components of FireFly v2 are arranged as illustrated in Fig. \ref{fig:bd}C. The blue blocks in Fig.\ref{fig:bd}C represent a series of input spike preprocessing modules, whereas the orange blocks signify a sequence of synaptic weight preprocessing modules. The purple blocks represent the post-processing modules for the partial sums and output spikes.
The input spike stream and synaptic weight stream pass through the preprocessing modules before reaching the spike computing engine. Subsequently, the spike computing engine generates a partial sums stream, which is then further processed by the post-processing modules to produce output spikes. We will provide a brief overview of these key components below.

\paragraph{Input Data Preprocessing}
The input spike preprocessing modules initially buffer and adjust the input spike stream's stream width via a FIFO and stream width adapter.
Subsequently, two sub-modules handle stream padding. One handles zero-padding based on the current convolution padding configurations, while the other focuses on memory coalescing to prevent bank conflicts.
After padding, the data stream enters the im2col unit sequentially but is then read out in a strided fashion to perform the im2col transformation. The im2col unit comprises $N$ memory banks, and a strided address generator generates $N$ conflict-free addresses for these banks to read the spike data. An $N$-port crossbar is responsible for routing the data output from the memory banks to their respective ports, thereby delivering the data stream to the spike computing engine.
The input weight stream undergoes a similar buffering and width adjustment procedure and then is pushed to the partial reuse FIFO, a component introduced in our prior version of FireFly\cite{li2023firefly}.

\paragraph{Clock Crossing in the Computing Engine}
The spike computing engine joins the spike stream from the im2col unit with the weight stream from the partial reuse FIFO and sends them to the fast clock region. The spike computing engine consists of two slow-to-fast gearboxes, functioning as parallel-to-serial converters. These gearboxes reduce the data elements by half while doubling the clock rate, maintaining the data bandwidth. The PEs(processing elements) of the spike computing engine operate at the fast clock region, performing matrix multiplications between the $V\times N\times S$ spike stream and the $M \times V$ weight stream and generating $M\times N\times S$ partial sum every $ K_h\times K_w \times \frac{C_i}{V}$ cycles. The partial sums are then gathered, aligned and sent back to the slow clock region.

\paragraph{Partial Sums Postprocessing}
The post-processing modules first flexibly process the partial sums, dealing with spike and non-spike cases, and then generate spikes through the neurodynamic unit.
The following pooling unit performs optional Maxpooling and AvgPooling or just bypasses the spike stream if pooling is not needed.
The residual connection module of the FireFly v2 performs the optional spike-element-wise residual connection between the shortcut spikes from the data loader and the calculated spike stream.
The spike accumulation module optionally counts spikes from the spike stream to record firing rates, a common operation in the last classification layer in SNN models.
The output spike stream flows back to the external memory map through the data saver and serves as the input spike stream for the subsequent layer.

In the following subsections, we will first fully elaborate on the design of the spike computing engine. Next, we will provide thorough details of how the spike computing engine and the partial sum merging unit cooperate to perform non-spike operations, which is essential in supporting direct encoding and multi-bit spike convolution. Additionally, we will delve into the two-phase design of the neurodynamics unit which can generate spikes across multiple time steps. Lastly, we will present the residual connection unit that supports spike-element-wise functions in various cases.

\subsection{Spatiotemporal Spiking Computing Engine}

\begin{figure}
  \centering
  \includegraphics[width=1.0\linewidth]{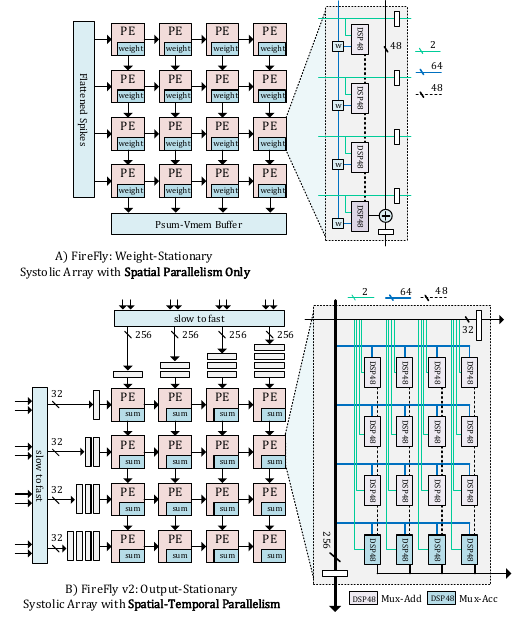}
  \caption{Comparision of the spike computing engine in FireFly v2 to its previous version of FireFly. A) Spike computing engine in FireFly supporting only spatial parallelism B) Spatiotemporal spike computing engine in FireFly v2.}
  \label{fig:systolic}
\end{figure}

The spike computing engine acts as the core of FireFly v2 since it is responsible for the heavy computing workload. We adopt the systolic architecture same as FireFly\cite{li2023firefly}, shown in Fig.\ref{fig:systolic}A, but with several distinctions:
1) FireFly employs a weight-stationary systolic array, whereas FireFly v2 implements an output-stationary systolic array. This choice is better aligned with our spatiotemporal dataflow requirements.
2) The systolic array in FireFly enables spatial parallelism across input channels, kernel sizes, and output channel dimensions. However, the parallelism in the kernel dimension imposes constraints on the convolution scheme, as FireFly exclusively supports $3\times 3$ convolutions. In contrast, FireFly v2 leverages spatiotemporal parallelism in input channels, output channels, pixel level, and the time-step dimension. We support various kinds of convolution configurations enabled by the flexible im2col unit.
3) In FireFly, the systolic array operates at a frequency of 300MHz, identical to the overall system clock frequency. FireFly v2 successfully decouples the slower fabric logic from the faster DSP unit. This method enables the spike computing engine to operate at 500-600MHz, doubling its performance capabilities compared to FireFly.

\begin{table}[]
  \begin{center}
    \caption{The Size of the Systolic Array and the PE}
    \label{enginesize}
  \begin{tabular}{c|c|c}
  \hline
  Notation & Description            & Determined by   \\ \hline\hline
  $SA_h$   & Systolic Array Height  & $M/4$           \\ \hline
  $SA_w$   & Systolic Array Width   & $N$             \\ \hline
  $PE_h$   & PE Height              & $V/4$           \\ \hline
  $PE_w$   & PE Width              & $S$             \\ \hline
  \end{tabular}
  \end{center}
\end{table}

Designing a high-performance systolic array is non-trivial. To bridge the gap between the operating frequency of DSP48E2 and its theoretical extreme frequency, we adopt the DSP double data rate technique as the Vitis AI DPU. We follow three key principles:
First, the circuits in the doubled-frequency domain should be well-decoupled from the circuits in the low-frequency domain.
Second, we avoid the use of LUTs in the doubled frequency domain, and instead, only use DSP48E2 and flip-flops.
Third, we avoid high-fanout nets and instead use simple and local connections between circuit components. This helps to minimize net delays and reduce congestion.

Before sending the spike sub tensor and weight sub tensor to the spiking computing engine, a slow-to-fast converter, or the gearbox, is utilized to facilitate communication between circuits operating at different frequencies. This gearbox, which operates at twice the frequency of the low-frequency domain, essentially functions as a multiplexer, selecting the data being transmitted from the low-frequency domain. The data being transmitted from the low-frequency domain has twice the data elements, but the data being transmitted to the doubled-frequency domain has twice the clock rate. As a result, the bandwidth at each side of the gearbox remains the same.

The spiking computing array is organized as a systolic array, with processing elements (PEs) arranged in a 2D fashion. To simplify the depiction, Fig.\ref{fig:systolic}B only illustrates a systolic array with $4\times 4$ PEs. The array employs an output stationary dataflow, with weight inputs flowing vertically from top to bottom and spike inputs flowing horizontally from left to right. Partial sums are stored in each PE and are collected once the accumulation process is complete.
Each processing element (PE) comprises several columns of DSP48E2s. To simplify the illustration, Fig.\ref{fig:systolic}B only depicts a single PE with four columns of DSP48E2s, where each column contains four DSP48E2s cascaded in a chain.
Similar to the previous version of FireFly\cite{li2023firefly}, a single DSP48E2 slice functions as a $2\times 4$ synaptic crossbar, receiving two 1-bit spike inputs and eight 8-bit weight inputs. The dedicated cascaded path of the DSP48E2 in the same column behaves like dendrites, integrating the synaptic current through the DSP48E2 adder chain. Each column of the processing element produces four 12-bit partial sums, utilizing the single instruction, multiple data (SIMD) feature of DSP48E2.
Within the same PE, DSP48E2s on the same row share the same weight inputs, while each DSP48E2 has its own spike inputs. In the illustrated example of a single PE with $4\times 4$ DSP48E2s, it receives $4\times 8\times 8=256$-bit weight inputs and $4\times 4\times 4=32$-bit spike inputs, generating $4\times 4\times 12=192$-bit partial sums. The weights and spikes are staged and then fed to the adjacent PEs, and the partial sums are collected after the accumulation process is completed.

The parallelism factors $M, V, N, S$ play a vital role in determining the dimensions of the systolic array. Specifically, Table.\ref{enginesize} outlines the relationship between these factors and the corresponding dimensions of the systolic array and PEs.
The height of the systolic array $SA_h$ is determined by $\frac{M}{4}$, where each column of DSP48E2 in one PE can compute 4 channels. The width of the systolic array $SA_w$ is directly equal to $N$. The height of the PE $PE_h$ is determined by $\frac{V}{4}$ due to the fact that the computing engine operates at a doubled frequency, and each DSP48E2 in one PE can integrate two synaptic currents. The width of the PE is determined by $S$.
It's worth noting that within a single Processing Element (PE), synaptic weights are broadcast to $S$ columns of DSP48E2 units. A critical consideration here is the choice of $S$, where a larger value would lead to a larger fan-out for synaptic weights, potentially failing step-up requirements. Conversely, opting for a smaller value of $S$ would elevate the consumption of flip-flops.
Based on experimental insights, we have determined the optimal value for $S$ to be 4. This empirical setting strikes a balance between managing fan-out effects and optimizing flip-flop usage.
We use different $SA_h$ and $SA_w$ in different FPGA devices with different amounts of resources.

The spike computing engine generates $M\times N\times S$ partial sums every $K_h\times K_w\times \frac{C_i}{V}$ cycles. Given that $K_h\times K_w\times \frac{C_i}{V}$ is larger than $N$ in most scenarios, we aggregate the partial sums $N$ in a group, align the partial sums from $\frac{M}{4}$ PE columns and use a cross clock region FIFO to transfer the $M\times N\times S$ partial sums back to slow clock region, shown in Fig.\ref{fig:bd}.

\subsection{Flexible Partial Sum Processing Unit}

\begin{figure}
  \centering
  \includegraphics[width=1.0\linewidth]{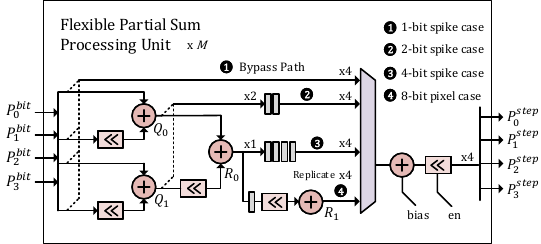}
  \caption{Flexible partial sum processing unit dealing with four spike or non-spike cases.}
  \label{fig:psum}
\end{figure}

Multi-bit spikes are decomposed into equivalent time steps using the same spike computing engine to compute the decomposed partial sums. The flexible partial processing unit shown in Fig.\ref{fig:psum} reconstructs the partial sums by shift-add logic.
The processing unit consists of $M$ identical sub-modules to processing $M$ channels of partial sums.
As stated in the previous subsection, $S$ is set to an empirical value of $4$, so each sub-module receives partial sums, namely $P_0, P_1, P_2, P_3$, of $4$ equivalent time steps,  shown in Fig.\ref{fig:psum}. The processing unit can handle 4 cases:

1) In cases where the input spike is binary, four partial sums are bypassed and directly sent to the next stage.

2) When dealing with a 2-bit input spike, two adjacent partial sums are shifted-merged, yielding $Q_0 = P_0 + (P_1<<1)$ and $Q_1 = P_2 + (P_3<<1)$. The processing unit waits for another round of the shift-merge process to collect 4 partial sums and send them to the next stage.

3) When dealing with a 4-bit input spike, all four partial sums are shifted-merged, yielding $R_0 = Q_0 + (Q_1<<2)$. The processing unit must wait for three additional rounds of the shift-merge process to gather four partial sums before transmitting them to the next stage.

4) When dealing with direct input coding where the input pixel is 8-bit, eight partial sums are shifted-merged, yielding $R_1 = R_0 + (R^{'}_0 << 4)$, in which $R^{'}_0$ is the previous round of $R_0$ temporarily stored in registers. In direct encoding, the convolution results of the static images are replicated $T$ times and sent to the neurodynamics unit. Therefore, we directly replicate $R_1$ for four times and send them to the next stage.

After the partial sums are shifted-merged, the bias is added, and the partial sums can be optionally left-shifted by 1 if the input spikes from the preceding layer are right-shifted by 1. The input precision of the partial sums is 12-bit. After being shifted and merged by the processing unit, the output precision of the partial sums is extended to 18-bit.

\subsection{Two-Phase Neurodynamics Unit}

\begin{figure}
  \centering
  \includegraphics[width=1.0\linewidth]{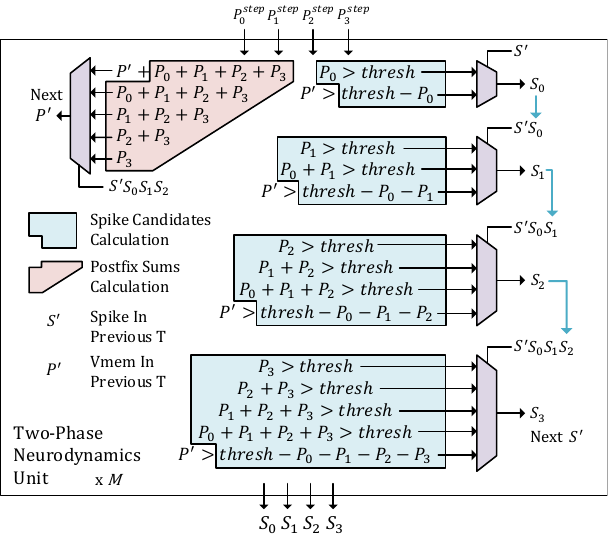}
  \caption{Neurodynamics unit. The blue-shaded logic performs independent calculations of spike candidates for each time step. The orange-shaded logic, on the other hand, handles the calculation of postfix sums for each time step. The purple multiplexer selects the spike candidates and postfix sums, generating both the output spikes and the residual membrane potential.}
  \label{fig:dynamic}
\end{figure}

After the partial sums are flexibly merged(or just bypassed in the 1-bit spike case), the partial sums flow into the neurodynamics unit.
In extreme cases, partial sums flowing into the neurodynamics unit are consecutively valid, the neurodynamics unit needs to process partial sums, generate spikes for all $S$ time steps and calculate the membrane potential for the next batch of time steps in just a single clock cycle.
The naive implementation of the neurodynamics unit is similar to the idea of the ripple carry adder, where each bit's sum depends on the carry generated by the previous bit, causing a serial carry propagation through the circuit. In the spike generation process, each spike and membrane potential depends on the previous spike and membrane potential, causing a series of data propagation through the circuit.
It is impossible to achieve timing closure at 250-300MHz with such high logic levels. 

To address this problem, we design a two-phase neurodynamics unit to tackle this problem. We take inspiration from the carry look-ahead adder, where the carry signals are precomputed for each bit, enabling parallel carry calculation and faster addition.
We decompose the spike generation process in two phases. Fig.\ref{fig:dynamic} shows the integrate-and-fire neurodynamics generation process.
In phase 1, we precompute the postfix sums of the partial sums of $S$ time steps and compare them to the threshold, yielding the spike candidates of each time step. In this phase, calculations are pipelined since they have no data dependency on previous time steps.
In phase 2, we select the spikes from these precomputed candidates based on spikes selected in previous time steps, as shown by blue arrows in Fig.\ref{fig:dynamic}. In this phase, timing closure can be satisfied since the accumulation of logic levels is only determined by the number of cascaded multiplexers and comparators.

The neurodynamics unit is statically reconfigurable to support integrate-and-fire\cite{abbott1999lapicque}, leaky-integrate-and-fire\cite{dayan2003theoretical} and residual membrane potential\cite{han2020rmp} neurons since most SNN models adopt a single neuron type across all layers. Designs of the neurodynamics unit of different neuron types share the same methodology.

\subsection{Residual Connection Unit}

The residual connection unit receives spikes of the current backbone from the flow-to-stream unit and receives spikes of the shortcut branch from the data loader. The residual connection unit performs IAND or ADD spike-element-wise function to the two spike streams.

To align with the 8-bit byte standard, the bit-width values of spikes from the shortcut branch are restricted to one, two, or four. 
The bit-width of spikes from the current backbone is one in most cases unless the optional average pooling datapath is selected. In that case, the bit-width increases to two.
FireFly v2 does not support residual connection after the backbone is downsampled by average pooling since such situations are not typical in most SNN models. Therefore, when performing residual connections, the bit-width of spikes from the current backbone is always one.

The residual connection unit contains dedicated logic of the IAND function for binary shortcut spikes and low-bit spike-element-wise ADD function for different shortcut spikes' bit-width.
The IAND function always produces binary spikes, which is the most hardware-friendly spike-element-wise function.
When performing the spike-element-wise ADD function, if the added results exceed the representation range of a two-bit integer, we can extend the added results to four bits or adopt the saturate-or-shift method to constrain the results back to two bits.
If the added results exceed the representation range of a four-bit integer, we will directly saturate the results to four bits.

\section{Implementation and Experiments}

\subsection{Experiments Setup}
Similar to its previous version, FireFly v2 targets FPGA edge devices to cut down the budget in real-world applications. FireFly v2 is mapped onto Ultra96v2, KV260 and ZCU104 evaluation boards with different $(M, V, N, S)$ configurations.

FireFly v2 is designed using SpinalHDL. The Verilog codes generated by the SpinalHDL compiler are synthesized and implemented in the Xilinx Vivado 2022.2. Power consumption estimates are provided by the reports of the Vivado Design Suite.
FireFly v2 is based on the Brain-Inspired Cognitive Engine (BrainCog)\cite{Zeng2023} and is another step toward the software–hardware codesigns for the BrainCog project (http://www.brain-cog.network/)\cite{braincogweb}. All the evaluated SNN models are trained using the BrainCog's infrastructures.

\subsection{Comparison with the Previous Version of FireFly}

\begin{table*}[]
  \centering
  \begin{threeparttable}[b]
  \caption{Comparison in Hardware Specifications between FireFly and FireFly v2}
  \label{fireflycmp}
  \begin{tabular}{c|c|c|c|c|c|c|c|c|c|c}
  \hline
                              & Device  & LUTs(K) & DSPs & B/URAM & Freq.   & Power & Array Size & Peak GOP/s & GOP/s/DSP & GOP/s/W \\ \hline\hline
  \multirow{2}{*}{FireFly}    & xczu3eg & 15      & 288  & 216/0  & 300     & 3.1\tnote{1}   & $144\times 16$     & 1382.4     & 4.8       & 445.94  \\ \cline{2-11} 
                              & xczu5ev & 42      & 1152 & 25/50  & 300     & 18.2  & $288\times 32$     & 5529.6     & 4.8       & 303.82  \\ \hline
  \multirow{3}{*}{FireFly v2} & xczu3eg & 23      & 256  & 103/0  & 300/600 & 6.2   & $16\times 16\times 4\times 4$  & \textbf{2457.6}     & \textbf{9.6}       & 396.39  \\ \cline{2-11} 
                              & xczu5ev & 26      & 512  & 87/8   & 250/500 & 4.9   & $16\times 16\times 8\times 4$  & \textbf{4096}       & \textbf{8}         & \textbf{835.92}  \\ \cline{2-11} 
                              & xczu7ev & 41      & 1024 & 160/8  & 250/500 & 13.5  & $32\times 16\times 8\times 4$  & \textbf{8192}       & \textbf{8}         & \textbf{606.81}  \\ \hline
  \end{tabular}
    \begin{tablenotes}
    \item[1] The reported power consumption in FireFly\cite{li2023firefly} is 2.55W, excluding the power consumption of the PS-side CPUs. To ensure a fair comparison, we use the total power consumption metric of the whole Zynq Ultrascale SoC in this table.
  \end{tablenotes}
\end{threeparttable}
  \end{table*}

  \begin{table*}[]
    \centering
    \begin{threeparttable}[b]
      \caption{Comparison with FireFly on Multiple SNN Models}
  \label{fireflybenchcomp}
    \begin{tabular}{ccllc|cccl|lccl|lccl}
    \hline
    \multicolumn{5}{c|}{Benchmark}                                                                                                 & \multicolumn{4}{c|}{FireFly(xczu3eg@300MHz)}                                                                    & \multicolumn{4}{c|}{FireFly v2(xczu3eg@600MHz)}                                                                 & \multicolumn{4}{c}{FireFly v2(xczu5ev@500MHz)}                                                                  \\ \hline\hline
    \multicolumn{1}{c|}{Net} & \multicolumn{1}{c|}{Dataset}     & \multicolumn{1}{l|}{FLOP}  & \multicolumn{1}{l|}{T} & Acc.  & \multicolumn{1}{c|}{us}   & \multicolumn{1}{c|}{W}   & \multicolumn{1}{c|}{FPS/W}  & \multicolumn{1}{c|}{GOP/s} & \multicolumn{1}{c|}{us}   & \multicolumn{1}{c|}{W}   & \multicolumn{1}{c|}{FPS/W}  & \multicolumn{1}{c|}{GOP/s} & \multicolumn{1}{c|}{us}   & \multicolumn{1}{c|}{W}   & \multicolumn{1}{c|}{FPS/W}   & \multicolumn{1}{c}{GOP/s} \\ \hline
    \multicolumn{1}{c|}{SNN5}  & \multicolumn{1}{c|}{MNIST}       & \multicolumn{1}{l|}{130M} & \multicolumn{1}{l|}{4} & 98.2 & \multicolumn{1}{c|}{491\tnote{1}}  & \multicolumn{1}{c|}{3.1} & \multicolumn{1}{c|}{656.9} & 1063.3                    & \multicolumn{1}{l|}{326}  & \multicolumn{1}{c|}{6.2} & \multicolumn{1}{c|}{494.7} & 1601.6                    & \multicolumn{1}{l|}{201}  & \multicolumn{1}{c|}{4.9} & \multicolumn{1}{c|}{1015.3} & 2597.6                    \\ \hline
    \multicolumn{1}{c|}{SNN7}  & \multicolumn{1}{c|}{CIFAR10}     & \multicolumn{1}{l|}{284M} & \multicolumn{1}{l|}{4} & 91.4 & \multicolumn{1}{c|}{1035\tnote{1}} & \multicolumn{1}{c|}{3.1} & \multicolumn{1}{c|}{311.6} & 1098.2                    & \multicolumn{1}{l|}{706}  & \multicolumn{1}{c|}{6.2} & \multicolumn{1}{c|}{228.4} & 1609.9                    & \multicolumn{1}{l|}{427}  & \multicolumn{1}{c|}{4.9} & \multicolumn{1}{c|}{477.9}  & 2661.9                    \\ \hline
    \multicolumn{1}{c|}{SNN11} & \multicolumn{1}{c|}{CIFAR100}    & \multicolumn{1}{l|}{586M} & \multicolumn{1}{l|}{4} & 64.3 & \multicolumn{1}{c|}{2128\tnote{1}} & \multicolumn{1}{c|}{3.1} & \multicolumn{1}{c|}{151.5} & 1101.7                    & \multicolumn{1}{l|}{1749} & \multicolumn{1}{c|}{6.2} & \multicolumn{1}{c|}{92.2}  & 1340.5                    & \multicolumn{1}{l|}{1057} & \multicolumn{1}{c|}{4.9} & \multicolumn{1}{c|}{193.1}  & 2218.1                    \\ \hline
    \multicolumn{1}{c|}{SNN9}  & \multicolumn{1}{c|}{DVS-G} & \multicolumn{1}{l|}{978M} & \multicolumn{1}{l|}{4} & 89.3 & \multicolumn{1}{c|}{3546\tnote{1}} & \multicolumn{1}{c|}{3.1} & \multicolumn{1}{c|}{90.9}  & 1103.7                    & \multicolumn{1}{l|}{1989} & \multicolumn{1}{c|}{6.2} & \multicolumn{1}{c|}{81.1}  & 1967.6                    & \multicolumn{1}{l|}{1281} & \multicolumn{1}{c|}{4.9} & \multicolumn{1}{c|}{159.3}  & 3055.2                    \\ \hline
    \end{tabular}
    \begin{tablenotes}
      \item[1,2,3,4] The reported inference latency in FireFly\cite{li2023firefly} do not include the direct coding layer, resulting the higher FPS/W and GOP/s metrics.
    \end{tablenotes}
  \end{threeparttable}
    \end{table*}

    \begin{table}[]
      \caption{Power Consumption of FireFly v2(xczu5ev) at Different Frequencies}
      \label{power}
      \begin{tabular}{c|c|c|c|c|c}
        \hline
        Frequency(MHz) & 300  & 350  & 400  & 450  & \textbf{500}  \\ \hline
        Power(W)     & 4.04 & 4.28 & 4.54 & 4.73 & \textbf{4.93} \\ \hline
        \end{tabular}
      \centering
      \end{table}

Table.\ref{fireflycmp} shows the comparison between FireFly v2 and its previous version FireFly in hardware specifications.
In terms of LUT consumption, FireFly v2 mapped on Ultra96 consumes slightly more LUTs than FireFly. This difference arises from the increased complexity of the overall architecture in FireFly v2.
However, FireFly v2 mapped on ZCU104 is roughly the same as FireFly since the proportion of the resource consumption taken up by the computing array becomes more significant as parallelism increases and FireFly v2 adopts a DSP-only and fabric-free spike computing engine.

In terms of DSP48E2 consumption, FireFly's DSP48E2 consumption aligns with multiples of 9 since FireFly seeks parallelism in kernel dimension by flattening the $3\times 3$ kernel window computation, while FireFly v2's DSP48E2 consumption aligns with multiples of 8 with each dimension in FireFly v2's parallelism being the power of two. Consequently, FireFly v2's DSP48E2 consumption is equivalent to $\frac{8}{9}$ of FireFly's consumption on the same device.
In terms of the DSP efficiency, power efficiency and throughput performance, FireFly v2 mapped on Ultra96 achieves the highest clock frequency of 600MHz and the highest peak DSP efficiency of $9.6$ GOP/s/DSP, which is doubled compared to FireFly. The DSP efficiency improvement of FireFly v2 compared to its previous version is primarily attributed to the increased clock frequency.
FireFly v2 mapped on KV260 achieves the highest peak power efficiency of $835.9$ GOP/s/W, maintaining a low power draw of 4.9W.
FireFly v2 mapped on ZCU104 achieves the highest peak throughput of $8192$ GOP/s.

FireFly v2 mapped on Ultra96 can reach 600MHz with PerformaceWithRemap implementation strategy set in Vivado Design Suite. However, this strategy induces higher power consumption. But still, FireFly v2 can achieve similar power efficiency compared to FireFly on the same Ultra96 device.

FireFly v2 mapped on KV260 cannot reach 600MHz even with PerformaceWithRemap strategy being enabled.  This limitation arises from the considerably inherent smaller CLB:DSP ratio of $\frac{93}{1}$ in KV260 in comparison to Ultra96 with CLB:DSP ratio of $\frac{196}{1}$. This translates to a higher likelihood of routing congestion that will cause degrade in frequency performance.
Nevertheless, FireFly v2 mapped on KV260 can reach timing closure at 500MHz and achieve excellent power efficiency when using PowerDefaultOpt implementation strategy. Since power consumption is tightly coupled to the clock frequency, we also run multiple experiments at lower frequencies using the same implementation strategy and find that FireFly v2 running at 500MHz achieves the best power efficiency, shown in Table.\ref{power}.
We also try a higher level of parallelism in KV260 since a $16\times 16\times 8\times 4$ configuration only utilizes 40\% of DSP48E2s. A $32\times 16\times 8\times 4$ configuration can meet timing closure at 400MHz. Note that in this configuration we halve the depth of the local weight cache depth and the FIFO size of the AXI DataMover, reducing the BRAM consumption to relieve the tight setup requirements.

FireFly v2 mapped on ZCU104 adopts a greater degree of parallelism to fully utilize the on-chip resources since ZCU104 is the largest FPGA device among the mentioned devices. FireFly v2 achieves $\times 2$ peak power efficiency and $\times 1.67$ peak throughput than FireFly on the same ZCU104 device.

We then compare FireFly v2 with our previous work on the same four SNN models, as initially reported in FireFly and displayed in Table.\ref{fireflybenchcomp}.
FireFly v2 mapped on xczu5ev shows $\times 1.54, \times 1.53$, $\times 1.27$ and $\times 1.76$ FPS/W improvements on the MNIST, CIFAR10, CIFAR100 and DVS-Gesture classification tasks respectively.
While FireFly v2 mapped on xczu3eg may not excel in terms of the FPS/W metric due to the power inefficiencies brought by the complex routings operating under a 600MHz clock frequency, it still exhibits a substantial improvement in inference latency and actual GOP/s performance on the same device compared with FireFly.
It's worth noting that the inference latency of our previous work, FireFly, does not include the direct coding layer, as FireFly does not support non-spike convolution. In contrast, the inference latency of FireFly v2, as presented in Table \ref{fireflybenchcomp}, is measured end-to-end.
The actual improvements of FireFly v2 in these metrics should be even higher.

One might also notice that the actual performance improvement is not directly proportional to the peak performance improvement shown in Table.\ref{fireflycmp}. This discrepancy is primarily due to FireFly v2 adopting a coarser parallelism granularity, which can be fully leveraged when processing input feature maps from larger datasets, such as ImageNet.
In FireFly, we specifically selected these four models with $3\times 3$ convolutional layers and max-pooling layers only, as FireFly is particularly well-suited for optimizing these types of layers.
FireFly adopts a fixed convolution configuration and a fully flattened parallelism scheme in the kernel dimension. The spike pixels in the same feature map are processed sequentially in an on-the-fly manner. This allows FireFly to handle small feature maps more effectively.
FireFly v2, on the other hand, supports general \texttt{torch.nn.Conv2d} operations but operates with a coarser granularity at the pixel level, as it supports pixel-level parallelism and can process $N$ spike pixels at a time.
As a result, we may not fully leverage its advantages when handling small feature maps on FireFly v2, especially when $N \geq W_o$.
Taking the CIFAR-10 or CIFAR-100 dataset as an example, the size of the feature map is initially only $32\times 32$. After three $2\times 2$ pooling operations, the size becomes $4\times 4$. When dealing with feature maps with a width or height smaller than $4$, several inefficiencies become apparent:
1) The explicit same-padding processing time becomes noticeable, as only $\frac{4\times 4}{6\times 6}=\frac{16}{36}$ spike pixels are valid.
2) When $N>4$, the redundant processing elements allocated for pixel parallelism remain idle.
3) Dealing with small feature maps reduces the opportunities for reusing kernel weights within the same set of feature maps, making parameter bandwidth a bottleneck.
These inefficiencies won't occur when dealing with large datasets such as ImageNet.
Despite the listing inefficiencies, FireFly v2 still achieves improvements in latency and efficiency on the same benchmarks compared with our previous work.

\subsection{Comparison with the DeepFire2}

\begin{table*}[]
  \centering
  \begin{threeparttable}[b]
    \caption{Comparison with DeepFire2 on Multiple SNN Models}
\label{cmpdeepfire}
  \begin{tabular}{c|c|c|c|c|c|c|c|c|c|c|c|c|c}
  \hline
                                                                                                        & Network      & Dataset  & FLOPs & T & Acc.  & us & DSP  & W & MHz & GOP/s    & DSP Eff.\tnote{1} & DSP Eff.\tnote{2} & GOP/s/W \\ \hline\hline
\multirow{5}{*}{\begin{tabular}[c]{@{}c@{}}FireFly v2\\ (xczu5ev)\end{tabular}} & CIFAR-Net\tnote{3}    & CIFAR10  & 2.58  & 4 & \textbf{93.6}  & 2997    & 512  & 4.9   & 500   & 3443  & 8            & \textbf{6.73}         & \textbf{702.74}  \\ \cline{2-14} 
                                                                                & CIFAR-Net\tnote{4}    & CIFAR100 & 4.08  & 4 & \textbf{74.7}  & 4502    & 512  & 4.9   & 500   & 3625  & 8            & \textbf{7.08}         & \textbf{739.81}  \\ \cline{2-14} 
                                                                                & SEW-ResNet34\tnote{5} & ImageNet & 7.34  & 8 & \textbf{67.3}  & 30613   & 512  & 4.9   & 500   & 1918  & 8            & 3.75         & 391.46  \\ \cline{2-14} 
                                                                                & SEW-ResNet34\tnote{6} & ImageNet & 7.34  & 8 & 62.4  & 20276   & 512  & 4.9   & 500   & 2896  & 8            & \textbf{5.66}         & \textbf{591.03}  \\ \cline{2-14} 
                                                                                & SEW-ResNet34\tnote{7} & ImageNet & 9.58  & 8 & 62.4  & 24696   & 512  & 4.9   & 500   & 3103  & 8            & \textbf{6.06}         & \textbf{633.33}  \\ \hline
\multirow{3}{*}{\begin{tabular}[c]{@{}c@{}}DeepFire2\\ (xcvu9p)\end{tabular}}   & VGG10-S      & CIFAR10  & 0.45  & 1 & 87.10 & 43      & 2050 & 20.1  & 550   & 10400 & 8.8          & 5.10         & 517.93  \\ \cline{2-14} 
                                                                                & VGG10-L      & CIFAR100 & 1.34  & 1 & 65.90 & 82      & 2881 & 29.8  & 500   & 15500 & 8            & 5.67         & 519.79  \\ \cline{2-14} 
                                                                                & VGG13-L      & ImageNet & 15.76 & 1 & 40.10 & 641     & 5400 & 47.2  & 450   & 21100 & 7.2          & 4.55         & 447.00  \\ \hline
  \end{tabular}
  \begin{tablenotes}
    \item[1,2] The first DSP Eff. represents the theoretical peak GOP/s/DSP metric, while the second represents the actual average GOP/s/DSP metric. 
    \item[3] 3x32x32-32c3-256c3-256c3-mp2-256c3-256c3-256c3-mp2-512c3-mp2-1024c3-ap-10
    \item[4] 3x32x32-64c3-256c3-256c3-mp2-256c3-512c3-512c3-mp2-512c3-mp2-1024c3-ap-10
    \item[5,6] The spike element-wise function of the first SEW-ResNet34 is ADD, while the second is IAND.
    \item[7] The spike element-wise function is IAND. The input image is resized to $3\times 256\times 256$ to align with the computation granularity of FireFly v2. 
  \end{tablenotes}
\end{threeparttable}
  \end{table*}

In FireFly, we've evaluated various systolic-array-based SNN accelerators. We won't repeat these comparisons, as FireFly has already shown superior performance compared to those prior studies\cite{neil2014minitaur}\cite{han2020hardware}\cite{zhang2019asynchronous}\cite{ju2020fpga}\cite{fang2020encoding}\cite{ye2022implementation}\cite{chen2022cerebron}\cite{gerlinghoff2021e3ne}\cite{panchapakesan2022syncnn}.
In this paper, we compare FireFly v2 with DeepFire\cite{aung2021deepfire} and DeepFire2\cite{aung2023deepfire2}, two recently published high-performance SNN accelerators also with DSP optimizations and operating at 450-600MHz high clock frequency.
DeepFire series targets large multi-die FPGA devices and adopts layer-wise mapping of the entire SNN models.
DeepFire2 achieves the highest clock frequency of 600MHz and throughput among all FPGA-based SNN implementations with deep pipelining.

It is important to figure out the experimental setup of DeepFire2 to ensure fair comparison.
Despite adopting distinct SNN model mapping schemes (Folded for FireFly, Unrolled for DeepFire), both series of accelerators utilize the same GOP/s metrics.
The FLOPS count for the SNN models is determined by calculating the FLOPS count of their equivalent ANN models using established tools like \texttt{ptflops}.
This same experimental setup enables a fair and meaningful comparison.
However, DeepFire2 did not provide information about their time step configuration in their experiments, a critical parameter that significantly impacts inference latency. Furthermore, it's important to note that DeepFire2 does not support any form of time-step aggregation or sparsity acceleration. Consequently, inference performance relies solely on the following factors: the total FLOPs of the model, simulation time step, clock frequency, and DSP usage, with the simulation time step being the only unknown variable.
Based on the metrics reported in their research, it can be inferred that DeepFire2 adopts a simulation time step of one, which explains the exceptionally low reported inference latency.
As the computation workload scales linearly with the time step, we quantify the computation workload as the product of FLOPs and the time step (FLOPs·T).

Moreover, in DeepFire series accelerators, SNN models for CIFAR-10, CIFAR-100, and ImageNet classification are meticulously crafted to ensure that the network parameters align seamlessly with the storage granularity of BRAM and URAM.
Although the performance of FireFly v2 does not strongly correlate with specific SNN models, we choose SNN models that align with the parallelism granularity of FireFly v2's architecture to ensure a fair comparison.

We have the following several key observations in Table.\ref{cmpdeepfire}.

1) Both FireFly v2 and DeepFire2 achieve significantly high clock frequencies, exceeding 400MHz. FireFly v2 exhibits consistent frequency performance as a standalone engine, while DeepFire2 experiences a sharp drop in frequency, dropping to 450MHz when deploying deep SNN models on large datasets.

2) DeepFire2 prioritizes inference latency over benchmark accuracy by adopting a $T=1$ SNN setup. In contrast, FireFly v2 targets SNN models capable of delivering high classification accuracy, particularly on more complex datasets such as CIFAR100 and ImageNet. The accuracy of 93.6\%, 74.7\%, and 67.3\% achieved on CIFAR10, CIFAR100, and ImageNet are closely aligned with the state-of-the-art performance in SNN algorithms. The remaining performance gap is primarily attributed to the quantization process, which could potentially be mitigated through the adoption of a quantization-aware training approach in the future.

3) FireFly v2 falls short in achieving the same level of GOP/s performance and inference latency as DeepFire2, since xcvu9p, the FPGA used by DeepFire2, is considerably larger than the edge devices we use.
However, it's noteworthy that FireFly v2 has $\times 1.32, \times 1.25, \times 1.33$ average GOP/s/DSP efficiency improvements, and $\times 1.35, \times 1.42, \times 1.42$ power efficiency improvements on CIFAR10, CIFAR100 and ImageNet classification tasks compared to DeepFire2.

4) The SNN models benchmarked by FireFly v2 compared to DeepFire2 are not only larger in terms of FLOPs (2.58 vs. 0.45 on the CIFAR10 task and 4.08 vs. 1.34 on the CIFAR100 task) but also more complex(ResNet compared to VGGNet on the ImageNet task). FireFly v2 also exhibits scalability, enabling support for even larger and deeper SNN models. In contrast, DeepFire2's solution is not scalable, as it's constrained by the FPGA on-chip resources, restricting the supported model complexity.

5) DeepFire2 relies on costly, large FPGA devices that may not be practical for deployment in embedded systems within edge scenarios. On the other hand, the KV260 device we employ is a commercially available and relatively affordable FPGA device. It ensures that the budget for constructing customized edge systems for real-world applications remains manageable.

It is worth noting that FireFly v2 will exhibit a higher performance when the convolutional configuration aligns with its computation granularity. For instance, in the case of SEW-ResNet34 with a $224\times 224$ image input, the resulting feature map widths of $14$ and $7$ do not align with the $\times 8$ pixel parallelism granularity of FireFly v2. However, when SEW-ResNet34 uses a $256\times 256$ image input, there is a notable improvement in efficiency, as the feature map size aligns with the $\times 8$ granularity.
It's also worth mentioning that relying on FLOP calculations of equivalent ANN models may not offer a fair measure of model capacity when running SNN models with multi-bit spikes. This is because the FLOP count of equivalent ANN models does not account for the bit-width of operands, resulting in an efficiency drop when benchmarking the SEW-ResNet34 network with the ADD function.

FireFly v2 and DeepFire2 both utilize similar DSP optimization techniques and operate at similar clock frequencies, resulting in comparable normalized inference efficiency.
FireFly v2's higher efficiency compared to DeepFire2 is primarily attributed to its systolic array consistently operating without idle states. At the same time, the remarkably low inference latency of DeepFire2 is primarily achieved through its use of single-time-step inference and extensive utilization of DSP48E2 resources.

\subsection{Discussion}

In our experiments, our primary focus is on comparing FireFly v2 accelerators with DeepFire2 since both of its previous versions have already outperformed most SNN FPGA accelerators in terms of latency and efficiency.

The excellent performance of our previous work FireFly is mainly attributed to utilizing the DSP48E2s to build a large synaptic crossbar circuit.
The improvements shown in FireFly v2 are attributed to the optimized spatiotemporal dataflow and the doubled clock frequency.
The remarkably low inference latency achieved by the DeepFire series primarily stems from their extensive utilization of DSP48E2 resources on a large FPGA device, combined with a relatively simplified SNN setup featuring a time step of one (though not explicitly reported in their paper).
Both the FireFly and DeepFire series show that building large-scale and high-speed parallel dedicated computing units can achieve significant inference acceleration even though the sparsity nature of SNNs is not considered.
However, we believe improving the performance of inference efficiency based on FPGA devices without sparsity acceleration becomes more and more challenging. The clock frequency of FireFly v2 and DeepFire2 is already close to the maximum supported frequency by Ultrascale+ FPGA devices.

We also recognize that it is impractical to develop an FPGA-based SNN accelerator that can match the power efficiency of ASIC designs, as FPGAs inherently trade some efficiency for reconfigurability.
While ASIC designers primarily prioritize power efficiency, FPGA designers concentrate on fully utilizing on-chip resources and tailoring their designs to match the characteristics of FPGA devices.
We have put this principle into practice by utilizing the DSP48E2 cells for synaptic operations and approaching the peak DSP48e2 frequency by decoupling the high-speed hard blocks from the low-speed fabric.

Another significant aspect of FireFly v2 is its advancement toward a general SNN-DPU solution, akin to Vitis-AI DPU—the ANN-DPU counterpart. The support for non-spike operations in FireFly v2 is crucial for end-to-end deployment without requiring algorithm modifications. This represents a milestone where SNN algorithmic research, such as DIET-SNN\cite{rathi2021diet} and SEW-ResNet\cite{fang2021deep}, can be directly deployed onto FireFly v2 with minimal impact on accuracy.
It is worth noting that SNN algorithmic research is still rapidly evolving in terms of encoding schemes\cite{qiu2023gated}, neuron types\cite{shen2023exploiting} and connection topologies\cite{hu2021advancing}.
While the development cycle for hardware used to be significantly longer than that for algorithm software, the current trend in FPGA-based agile hardware development can now expedite the process and provide timely support for the latest algorithmic advancements.

\section{Conclusions}

FireFly v2 exhibits significant improvements over our initial version of FireFly.
It takes a significant step forward in advancing hardware support for current SNN algorithm developments by supporting non-spike operation, which presents an obstacle in the end-to-end deployment onto existing specialized SNN hardware.
The spatiotemporal dataflow enables the processing of incoming spikes and the generation of output spikes on the fly.
Additionally, the double data rate technique enables the DSP48E2 systolic array to operate at a clock frequency of 500-600MHz, which is twice as fast as our previous version of FireFly.
In this work, our focus remains on targeting commercially available and affordable embedded FPGA edge devices for use in edge scenarios.
In the future, we will continue to develop SNN hardware infrastructures that can not only operate at higher speeds but also offer timely support for advancements in SNN algorithms, enabling higher-performance SNN software and hardware co-design.

\bibliographystyle{IEEEtran}
\bibliography{myIEEE,reference}
\end{document}